\documentclass[sn-mathphys,Numbered]{sn-jnl}

\usepackage{graphicx}%
\usepackage{multirow}%
\usepackage{amsmath,amssymb,amsfonts}%
\usepackage{amsthm}%
\usepackage{mathrsfs}%
\usepackage[title]{appendix}%
\usepackage{xcolor}%
\usepackage{textcomp}%
\usepackage{manyfoot}%
\usepackage{booktabs}%
\usepackage{algorithm}%
\usepackage{algorithmicx}%
\usepackage{algpseudocode}%
\usepackage{listings}%
\usepackage{subcaption}
\usepackage{arydshln}

\theoremstyle{thmstyleone}%
\theoremstyle{thmstyletwo}%

\theoremstyle{thmstylethree}%

\begin{document}

\title[Group Activity Recognition using Unreliable Tracked Pose]{Group Activity Recognition using Unreliable Tracked Pose}


\author*[1]{\fnm{Haritha} \sur{Thilakarathne}}\email{p.thilakarathne@latrobe.edu.au}

\author[1]{\fnm{Aiden} \sur{Nibali}}\email{a.nibali@latrobe.edu.au}

\author[1]{\fnm{Zhen} \sur{He}}\email{z.he@latrobe.edu.au}

\author[1,2]{\fnm{Stuart} \sur{Morgan}}\email{stuart.morgan@sportaus.gov.au}

\affil*[1]{\orgdiv{Department of Computer Science \& Information Technology}, \orgname{La Trobe University}, \orgaddress{\street{Plenty Road \&, Kingsbury Dr}, \city{Bundoora}, \postcode{3086}, \state{VIC}, \country{Australia}}}

\affil[2]{\orgdiv{Sport Strategy \& Investment}, \orgname{Australian Institute of Sport}, \orgaddress{\street{Leverrier St}, \city{Bruce}, \postcode{2617}, \state{ACT}, \country{Australia}}}


\abstract{Group activity recognition in video is a complex task due to the need for a model to recognise the actions of all individuals in the video and their complex interactions.  Recent studies propose that optimal performance is achieved by individually tracking each person and subsequently inputting the sequence of poses or cropped images/optical flow into a model. This helps the model to recognise what actions each person is performing before they are merged to arrive at the group action class. However, all previous models are highly reliant on high quality tracking and  have only been evaluated using ground truth tracking information. In practice it is almost impossible to achieve highly reliable tracking information for all individuals in a group activity video. We introduce an innovative deep learning-based group activity recognition approach called Rendered Pose based Group Activity Recognition System (RePGARS) which is designed to be tolerant of unreliable tracking and pose information.  Experimental results confirm that RePGARS outperforms all existing group activity recognition algorithms tested which do not use ground truth detection and tracking information.}


\maketitle

\section{Introduction}
\label{sec:Introduction}

In the rapidly evolving landscape of computer vision and video analytics, recognising group activities in complex scenarios such as sporting events poses a formidable challenge. Traditional methods predominantly operate directly on video pixel data \cite{Ibrahim_2016_hierarchical_deep, CERN, social_scene} or rely on accurate human tracking and pose detection \cite{Perez2020_SkeletonBased_GIRN, Gavrilyuk2020_ActorTransformer, thilakarathne_2022_pogars} to discern individual entities within a scene. In most sports analysis scenarios, manual tracking and pose annotations of players are not readily available. Therefore, the utilisation of pose detection and tracking algorithms \cite{zheng2023_deep_pose_estimation} has become a prevalent method for extracting meaningful information from video data.

The majority of current group activity recognition approaches  \cite{Lu2019_spatioTempAtt, thilakarathne_2022_pogars} that use pose information assume detection and tracking data are accurate. While ground truth information generally ensures superior accuracy, obtaining such data in complex. This is especially true for uncontrolled settings such as sporting events, where hardware-based tracking systems are unavailable. In such scenarios, the most reliable option is post-hoc manual annotation, but the high cost of this solution limits its scalability. Consequently, ground truth tracked pose annotations typically do not accompany real-world sports footage. An alternative is to utilise real-time detection and tracking information extracted from videos by an algorithm. However, real-time detection and tracking is prone to defective pose estimates and broken tracks, thereby hindering the accurate prediction of group activities by existing approaches assuming perfect tracked pose. It is therefore important for any models making use of this data to be robust to such imperfections.

Most high performing existing pose-based group activity recognition approaches use the late fusion approach \cite{Ibrahim_2016_hierarchical_deep, thilakarathne_2022_pogars} where each person's tracked pose is fed separately into a model. This allows the model to observe the motion of all the joints of a person in sequence, in order to infer the action the person is performing. However, when a track is broken and a new one started then the model is unable to observe the continuous motion of the person being tracked. This creation of many small broken tracks results in the model being fed the pose of many separate short partial actions. It significantly degrades the model's ability to gain a high level understanding of the activity being performed by the group of people. Early person fusion approaches (where the pose from each person is first combined spatially before combined temporally) is also susceptible to degradation from poor tracked poses because the unreliable tracks results in faulty pose keypoint estimations. 

In this study we propose \textbf{Rendered Pose based Group Activity Recognition System (RePGARS)}, an approach designed to work well for group activity recognition in videos in the presence of unreliable tracked pose. Unlike prior approaches where broken tracks result in separated input, RePGARS renders the pose into images, where individuals are assigned different colours. In this setting, a broken track just manifests as a new colour being assigned to the same person, the model is still able to track the continuous motion of the person using a pre-trained 3D CNN. We also feed the original RGB image as input to the model to give it more context information to recover from broken tracks and incorrect pose estimations.

In our experiments, RePGARS outperforms all existing group activity recognition algorithms that we tested which do not use ground truth detection and tracking information. RePGARS outperforms POGARS \cite{thilakarathne_2022_pogars} by 12.8\% for the volleyball dataset \cite{Ibrahim_2016_hierarchical_deep} and 26.3\% for the Australian Netball Video dataset when trained and evaluated using unreliable tracked pose information as the input. These results are obtained solely through real-time pose detection and tracking information generated using the \textit{OpenPifPaf} algorithm \cite{Kreiss2021_openpifpaf}. 

Performance of RePGARS degrades by only 1.1\% for the volleyball dataset when unreliable tracked pose is used instead of ground truth tracking. In contrast, POGARS performance degrades by 14.3\% when unreliable tracked pose information is used. This highlights the advantage of using rendered pose as input to our RePGARS method in accurately recognising group activities with less reliable detection and tracking information compared to the keypoint based pose representation used by POGARS.

Our contributions can be summarised as follows: 

\begin{itemize}
\item We introduce RePGARS, an innovative system for recognising group activities which operates by utilising real-time individual detection and tracking data as its primary input. Our experiments indicate that RePGARS performs outperforms existing methods in scenarios where reliable detection and tracking information is not available.

\item Through a series of experimental results, we demonstrate the effectiveness of a rendered pose-based approach in capturing the intricate spatial and temporal relationships inherent in group activities in sports videos.

\item We created the Australian Netball Video dataset (ANV dataset) which contains long untrimmed netball game videos with group event annotations to test the effectiveness of RePGARS in identifying group activities in sporting scenarios.  

\end{itemize}

\section{Related work}
\label{sec:Related-work}
In this section we review existing work in the area of group activity recognition and the use of deep learning techniques to solve the problem.

Group activity recognition involves the analysis of multi-person behavioural and interaction dynamics to interpret instances of group activities. Early efforts in this field predominantly rely on models that acquire information from video frames through hand-crafted features. Notable methodologies, including hierarchical graphical models \cite{Lan2012_discriminativeLatentModels, Amer2014_hirf} and dynamic Bayesian networks \cite{Zhu2013_contextAware}, have been employed to interpret group activities in video settings.

Choi et al. \cite{Choi2009_colllectiveActivity} introduced a spatio-temporal feature descriptor based on shape context \cite{Belongie2002_shapeMatching} to analyse group activities involving people. Hierarchical models, exemplified by \cite{Lan2012_discriminativeLatentModels}, interpret group activities by representing the actions of individual participants and their interactions through a tree-like graphical descriptor.

However, employing handcrafted feature engineering approaches for group activity recognition presents certain drawbacks, including domain dependency, substantial computational costs, and a resulting decrease in accuracy.

Recent advancements in group activity recognition models have seen a transition towards the integration of deep neural networks \cite{wu2021comprehensive}. These models incorporate video pixel data as well as data derived from human tracking and pose detection for recognising the spatio-temporal dynamics of individuals within a scene. 

Ibrahim et al. \cite{Ibrahim_2016_hierarchical_deep} proposed a two-level RNN based hierarchical method for group activity recognition which consists of two LSTM based sub modules. The first module encodes the individual player level actions and their temporal dynamics while a second one fuses outputs from the first module for providing temporal dynamics of the group activity. Inspired by this work, using spatial features have become the most popular approach for group activity recognition \cite{Ibrahim_2016_hierarchical_deep,Ramanathan_2016_keyperson,Tsunoda2017_footballAction,Qi2019_stagNet}. These models assume ground truth bounding boxes of the individuals and 2D CNNs to extract spatial features of them, most of them don't consider the spatial position of the individuals. In contrast to the above existing work, our proposed model uses person level pose and spatial location information (position tracklets) for predicting the group activity since the fusion of pose and location information with RGB data results a dense information representation. Moreover, instead of ground truth bounding boxes we utilise real-time detection and tracking information.

Researchers have developed different methods to utilise pose information to learn the spatio-temporal dynamics of the individuals in videos. Azar et al.'s  multi-stream CNN framework \cite{Azar2018_multistream} use pose heatmaps alongside with other input modalities including RGB frame, optical flow and warped optical flow. \cite{thilakarathne_2022_pogars} uses ground truth detection and tracking information in videos to identify individuals in videos and subsequently use those bounding boxes to extract pose keypoints of the players in Volleyball videos. The numerical representations of the extracted pose skeleton data have been used as the input for the group activity classification model. Following a similar approach for pose representation, \cite{Gavrilyuk2020_ActorTransformer, Perez2020_SkeletonBased_GIRN} use reliable ground truth bounding boxes to extract pose keypoints of the individuals in videos. Our approach abstains from relying on ground truth detection and tracking data, given its challenging availability in real-world human activity analysis scenarios.

\section{Australian Netball Video Dataset}
\label{sec:anv-dataset}

During the study, we observed a gap in large video datasets within the domain of activity recognition. Most existing datasets focus on individual actions performed by a limited number of people, leaving a notable gap in the availability of datasets featuring group activities (especially those related to sports). Although datasets such as \textit{Kinetics} \cite{kay2017_kinetics}, \textit{THUMOS14} \cite{THUMOS14}, \textit{UCF101} \cite{soomro2012_ucf101}, \textit{HMDB51} \cite{Kuehne_HMDB} and \textit{Sports-1M} \cite{large_scale_videoclassi} are prominent datasets used for training activity recognition models, they predominantly emphasise individual actions or activities performed within specific contexts. \textit{Volleyball dataset} \cite{Ibrahim_2016_hierarchical_deep} and the recently introduced \textit{NBA dataset for Sports Video Analysis (NSVA)} \cite{dew2022_NSVA} are two of the few publicly available sports video datasets which specifically focus on group activity recognition. This limitation hampers the development and evaluation of robust group activity recognition models. 

\begin{figure}[h]%
\centering
\includegraphics[width=1.0\textwidth]{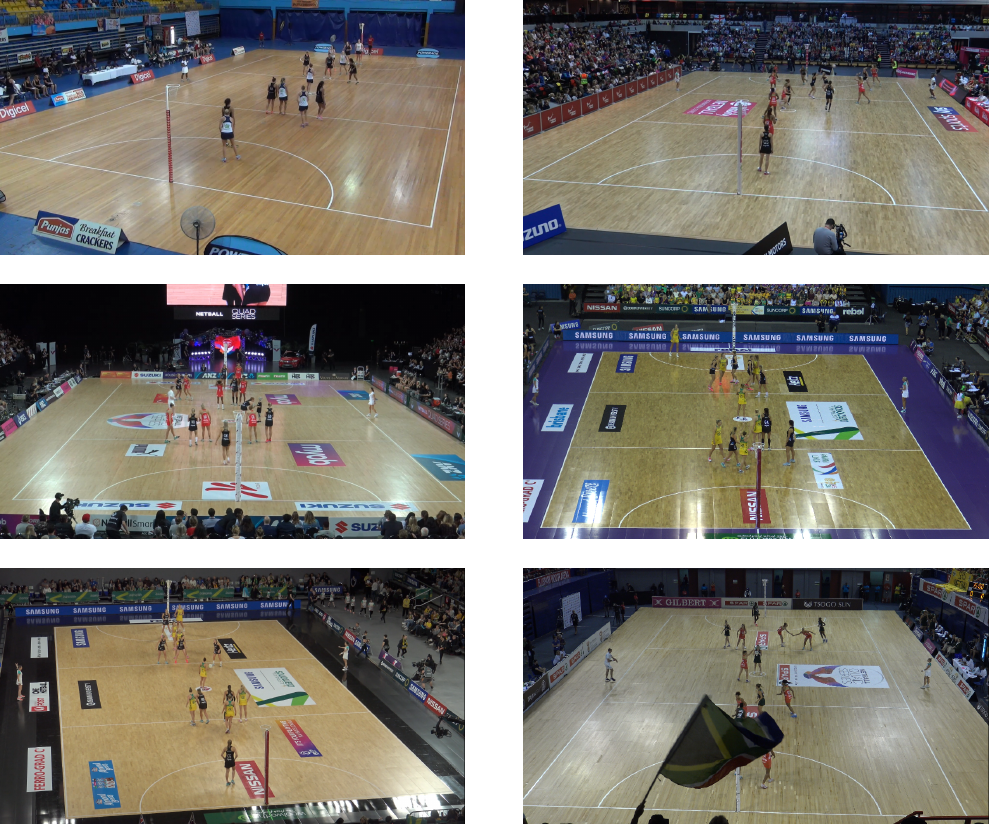}
\caption{Sample video frames from varied venues in the Australian Netball Video dataset.}
\label{f:netball-sample-video-frames}
\end{figure}

Addressing the gap in sports video datasets, we introduce a novel sports video dataset: the \textbf{Australian  Netball Video (ANV) dataset}, comprising footage from 11 netball matches played by Australian Diamonds Netball team. The dataset includes 17 untrimmed videos, each spanning approximately 15 minutes in duration. Each video covers one quarter of a netball match, showing the game in detail. The recordings have been captured across diverse indoor venues, employing a mounted camera within the arena. The camera positioning and the recording angle is different in each match. \hyperref[f:netball-sample-video-frames]{Figure~\ref*{f:netball-sample-video-frames}} illustrates sample frames from 6 varied venues where the netball games has been recorded for the dataset. One of the important features of this dataset is its high resolution; all footage was captured in 4K resolution (3840 x 2160 pixels), ensuring a clear and detailed visual representation of the game-play. 

\begin{figure}[h]%
\centering
\includegraphics[width=1.0\textwidth]{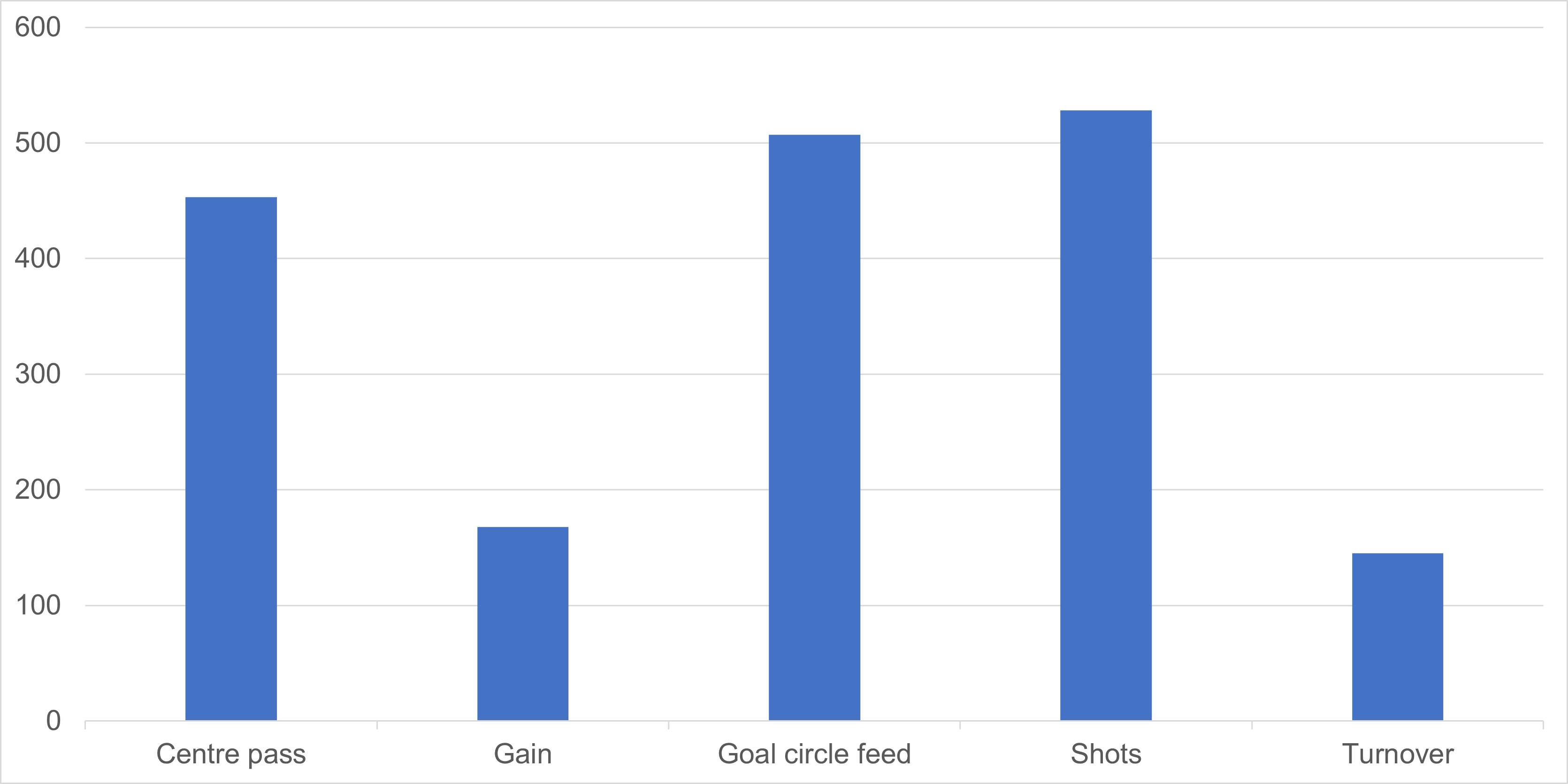}
\caption{Distribution of event classes annotated in the Australian Netball video dataset.}
\label{f:netball-activity-distribution}
\end{figure}

1801 instantaneous events within the netball videos were manually annotated, precisely marking the exact time when each event occurred. These annotations were carried out by expert netball game analysts affiliated with the Netball Australia. The annotated event classes include \textit{shot, centre pass, goal circle feed, turnover}, and \textit{gain} events in the context of netball game-play. The distribution of event classes annotated in the dataset is plotted in \hyperref[f:netball-activity-distribution]{figure~\ref*{f:netball-activity-distribution}}. A high level overview of these key events (according to \cite{Netball_rules}) is as follows: 

\begin{itemize}
  \item \textit{Centre pass:} A centre pass occurs at the beginning of each quarter and after every goal. It's when a player from the centre position passes the ball into the game, initiating the next phase of play. The player passing the ball must start inside the center circle. 
  
  \item \textit{Goal circle feed:} A goal circle feed happens when a player passes the ball to a teammate inside the shooting circle.

  \item \textit{Shot:} A shot is the attempt to score a goal. It takes place when a player in the shooting circle throws the ball towards the net, trying to score by getting the ball through the opponent's goalpost.

  \item \textit{Gain:} A gain refers to the successful retrieval of the ball by a defending player.

  \item \textit{Turnover:} A turnover takes place when the attacking team loses possession of the ball to the defending team.  
\end{itemize}

We believe that the ANV dataset is a useful asset for developing solutions to different sports video analysis tasks, including group activity recognition and instantaneous event detection. This study is focused on  using the dataset for recognising group activities, showcasing its practicality in computer vision based sports analysis.

\section{Challenges with Unreliable Tracking Data for Group Activity Recognition}
\label{sec:problems-of-unreliable-tracking}

Real-time pose detection and tracking algorithms are prone to error due to the inherent complexity and dynamic nature of human movements in interaction-heavy activities such as sports. Sports involve rapid movements, occlusions, complex interactions between players, varying lighting conditions, and unpredictable motion, all of which contribute to the ambiguity of tracking and pose estimation. This results in frequent broken tracks, identity switches and poor pose keypoint estimation. 

\begin{figure}[h]%
\centering
\includegraphics[width=1.0\textwidth]{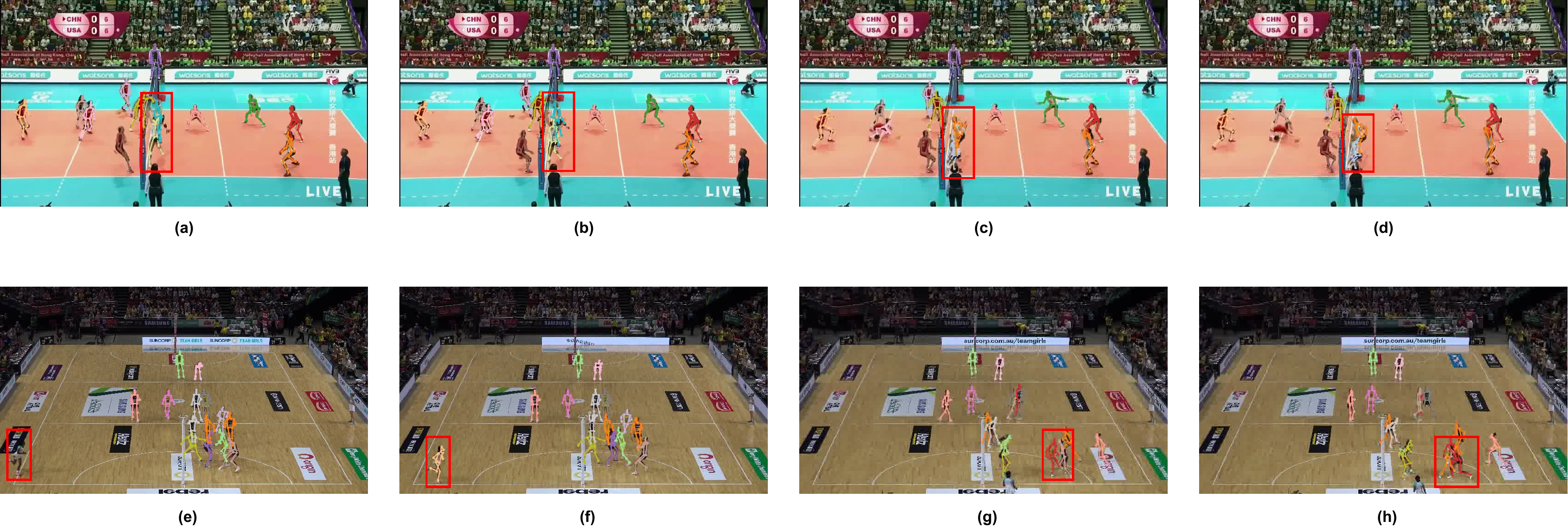}
\caption{Examples of unreliable tracking scenarios.}
\label{f:track-fails}
\end{figure}

Bottom-up pose detection and tracking approaches \cite{zheng2023_deep_pose_estimation} typically exhibit better scalability and efficiency in real-time applications \cite{jin2017_towards_posetracking}. However, we observe that bottom-up approaches often struggle to accurately interpret and handle dynamic elements, leading to errors in both tracking and pose estimation. \hyperref[f:track-fails]{figure~\ref*{f:track-fails}} visualises frame-wise examples of failed pose tracking. These visualisations were generated by applying \textit{OpenPifPaf} pose detection and tracking algorithm \cite{Kreiss2021_openpifpaf} to two video snippets sourced from the volleyball dataset \cite{Ibrahim_2016_hierarchical_deep} and the ANV dataset. Within the visualisations, the rendered pose detection of individuals are superimposed onto each frame, with distinct colours used to differentiate between detected individuals. The colour of the pose represents the identity of the track, and should remain consistent for the same player across frames.

In frames \textit{(a)} and \textit{(b)}, the red bounding box contains two players, with cyan and yellow colouring respectively. These same players are erroneously represented as entirely new individuals in frames \textit{(c)} and \textit{(d)}, now rendered in orange and grey. This discrepancy exemplifies a scenario in which the detection algorithm fails to consistently track individuals across frames. Moving to frames \textit{(e)} and \textit{(f)}, the red bounding box encapsulates a netball umpire moving across the court, leading to a false detection of an individual, erroneously identified as a netball player. Even in frames \textit{(g)} and \textit{(h)}, where the players inside the red bounding box appear identical, the detection algorithm assigns different identifiers to them. 

As visualised in \hyperref[f:track-fails]{figure~\ref*{f:track-fails}}, broken tracks with different track identifiers for the same individual is a common issue with real-time tracking and detection algorithm predictions. Such errors make it very difficult for pose based activity recognition algorithms such as POGARS \cite{thilakarathne_2022_pogars} to correctly determine the action a player is performing. 

\begin{figure}[h]%
    \centering
    \begin{subfigure}[b]{0.48\textwidth}
        \includegraphics[width=\textwidth]{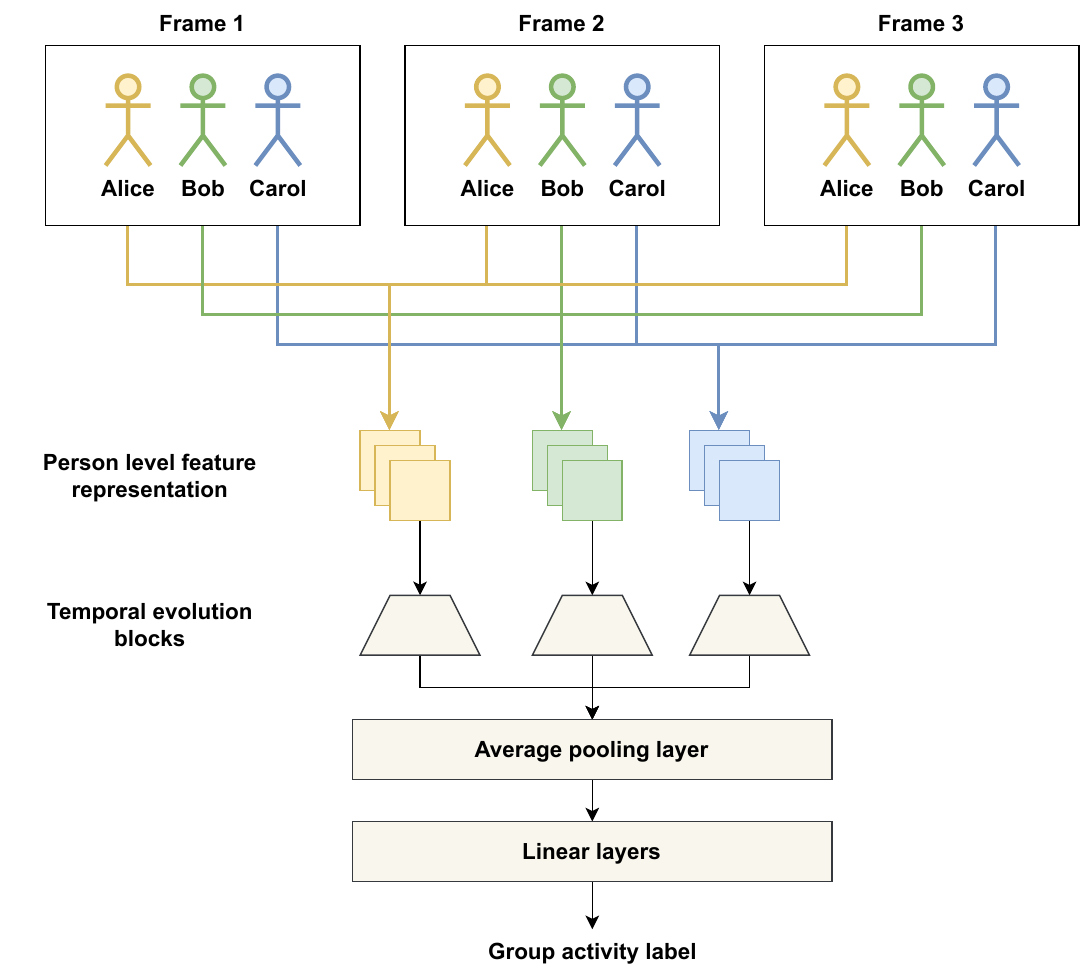}
        \caption{Reliable tracking scenario}
        \label{subfig:late-fusion-1}
    \end{subfigure}
    \hfill
    \begin{subfigure}[b]{0.48\textwidth}
        \includegraphics[width=\textwidth]{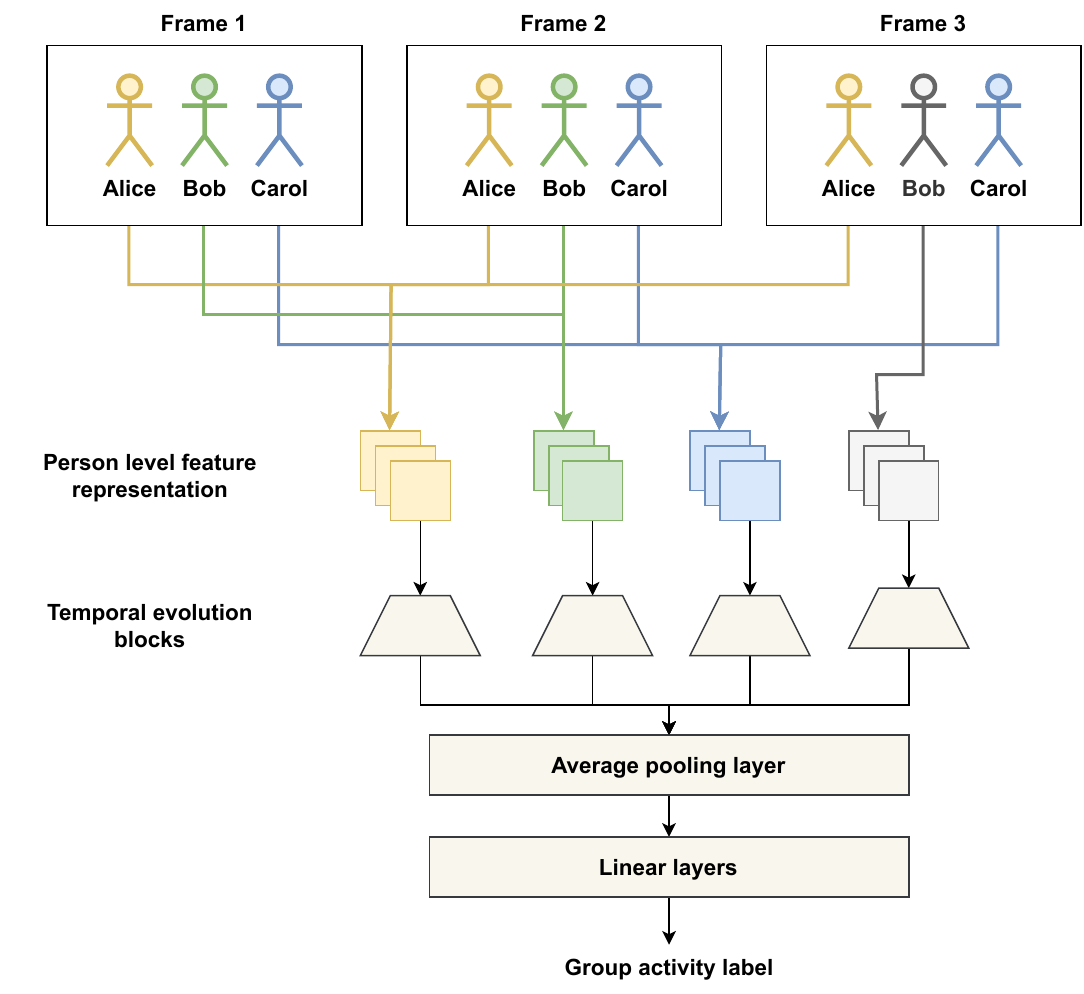}
        \caption{Unreliable tracking scenario}
        \label{subfig:late-fusion-2}
    \end{subfigure}
    \caption{Late person fusion approach. Similar colour individuals in each frame belongs to a single tracking identifier.}
    \label{f:late-fusion-problem}
\end{figure}

Pose information can be passed into models via either \textit{early person-level fusion} or \textit{late person-level fusion} as was described in \cite{thilakarathne_2022_pogars}. Poor tracking information effects early and late person-level fusion very differently. In early person-level fusion, all pose information are combined using the addition operation thereby effectively removing any identity information.

In late person fusion the temporal evolution of each person is first modeled individually before the learned features are fused together as shown in \hyperref[f:late-fusion-problem]{Figure~\ref*{f:late-fusion-problem}}. The \textit{reliable tracking scenario} represents flawless tracking where each individual assigned the correct tracking identifier throughout the temporal span of the video. In the \textit{unreliable tracking scenario}, \textit{Bob} has been erroneously identified as a new individual in \textit{frame 3} which demonstrates unreliable tracking. The consequence of this error is the model will find it very difficult to infer the action being performed by \textit{Bob} since it does not see a continuous sequence of joint movements for \textit{Bob}. As mentioned above such errors in tracking occurs often, thus making it extremely difficult for the model to gain a good high level understanding of the actions performed by all the individuals in the video. 

\begin{figure}[h]%
    \centering
    \begin{subfigure}[b]{0.48\textwidth}
        \includegraphics[width=\textwidth]{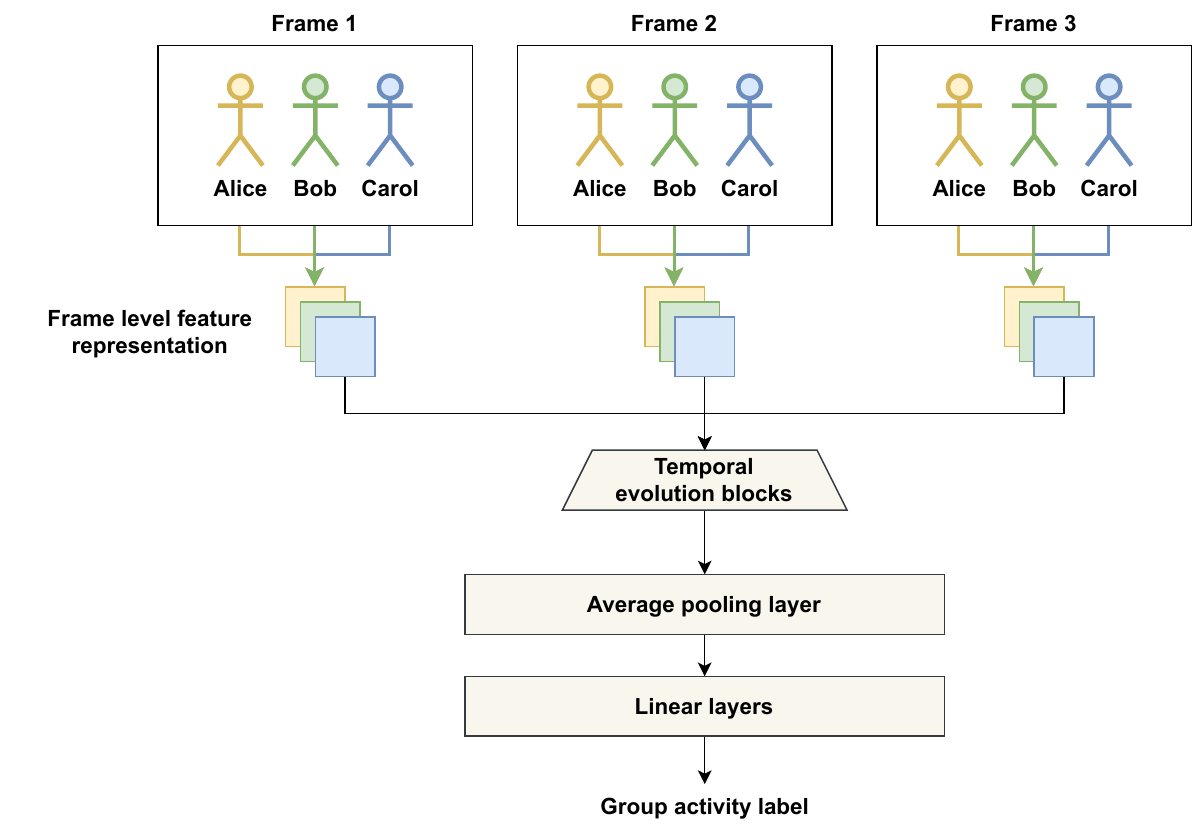}
        \caption{Reliable tracking scenario}
        \label{subfig:early-fusion-1}
    \end{subfigure}
    \hfill
    \begin{subfigure}[b]{0.48\textwidth}
        \includegraphics[width=\textwidth]{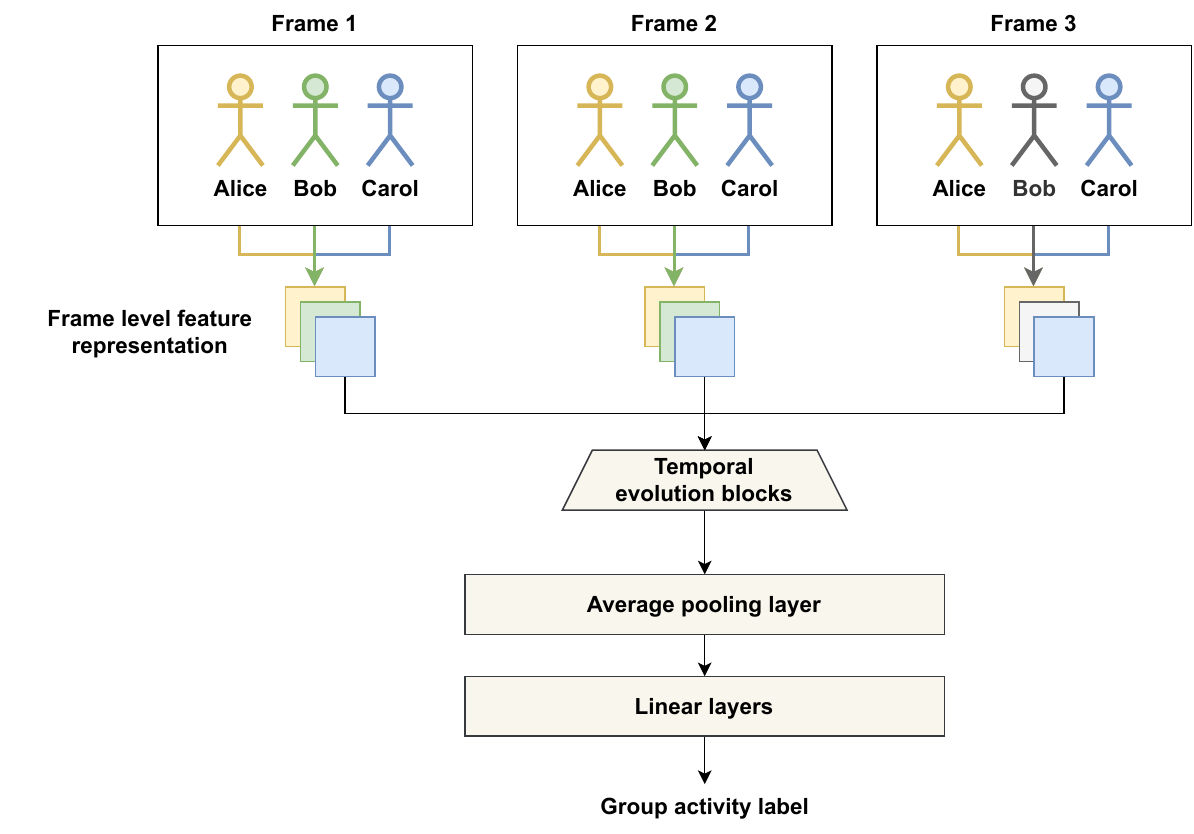}
        \caption{Unreliable tracking scenario}
        \label{subfig:early-fusion-2}
    \end{subfigure}
    \caption{Early person fusion approach. Similar colour individuals in each frame belongs to a single tracking identifier.}
    \label{f:early-fusion-problem}
\end{figure}

\hyperref[f:early-fusion-problem]{Figure~\ref*{f:early-fusion-problem}} illustrates early person-level fusion performed for the detection of \textit{Alice, Bob} and \textit{Carol} in three video frames. The \textit{reliable tracking scenario} represents flawless tracking where each individual is having same tracking identifier throughout the temporal span of the video. In the \textit{unreliable tracking scenario}, \textit{Bob} has been erroneously identified as a new individual in \textit{frame 3} which demonstrates unreliable tracking.
Early fusion removes person-level feature independence before modeling temporal evolution by fusing the tracked pose features of all people before input into the module that captures temporal dynamics. 

While early fusion effectively ignores the tracking information and therefore should suffer less when faced with broken tracks, it still performs worse than late fusion in general. The reason is early fusion does not see a sequence of joint movements for each person as a separate sequence, making it much more challenging to determine what action each person is performing. Experiments show person-wise spatio-temporal feature learning is essential for achieving higher classification accuracy (presented later in \hyperref[tab:pogars-vs-repgars]{table~\ref*{tab:pogars-vs-repgars}})

Real-time pose detection and tracking methods encounter challenges in capturing accurate human pose information within the context of sports videos. This highlights the need for developing group activity recognition models designed to be robust to errors caused by unreliable detection and tracking information whilst still learning person-level temporal dynamics in a scene.

\section{Rendered Pose based Group Activity Recognition System (RePGARS)}
\label{sec:repgars}

Prevailing deep learning based group activity recognition methods can be separated into two broad categories based on their input modalities. The first involves utilising RGB input and/or extracted optical flow data to learn spatio-temporal dynamics of individuals \cite{Ibrahim_2016_hierarchical_deep, Azar2018_multistream, Azar2019_convoRelationalMachine}, while the second uses pose and track information of the individuals in the video \cite{Lu2019_spatioTempAtt, Perez2020_SkeletonBased_GIRN, thilakarathne_2022_pogars}. Feeding the tracked pose information for each person separately to the model, makes it much easier for the model to predict what action each person is performing. This can then be combined to determine the group action. However, these methods are usually trained and tested using ground truth tracking information, which is not practical in most of the real-world settings.  

Though RGB input based methods are relatively straightforward and easier to use in practical applications, they fall short in capturing the intricate movement dynamics exhibited by individuals in the video footage \cite{franco2020multimodal}. Moreover, their sole reliance on colour information may result in spurious correlations when the training dataset is small. This is because the model has not been constrained in any way to encourage a focus on human movements. These methods are also sensitive to variations in lighting conditions, differences in venues, background and uniform players wearing impacting their generalisation across different environments, especially in low-light or night scenes. On the other hand, the accuracy of the pose information based approaches heavily depend on the efficacy of the pose detection and tracking algorithms employed. 

Pose detection and tracking in dynamic and crowded environments comes with several challenges. Individuals in crowded scenes can be partially or completely occluded by other objects or individuals making it difficult to maintain continuous tracking and reliable pose keypoint estimation.

In response to these limitations, we designed RePGARS to represent pose in a way that is more forgiving of broken tracks. Existing methods feed the pose of each person into the model separately, representing each pose as a collection of keypoints. As an alternative, we propose rendering each person's pose into a single image, denoting person identities with unique colours (\hyperref[f:rendered-pose-sample]{Figure~\ref*{f:rendered-pose-sample}}). We remove the background so the model can focus its attention on the pose itself. Since the colour assigned to each person remains temporally consistent when tracking is correct, the model can easily associate the pose of each person across time. However, when the tracking is wrong and the colour of a person's pose changes as a result, the model has the ability to learn its own tracking by ignoring or placing less weight on colour of the pose. This potential to correct broken tracks is possible because the model is trained in an end-to-end manner using the label for the group activity. In contrast, a broken track in the case of late-person fusion will result in broken fragments of the tracked pose for the same person being feed into the model separately. It is much harder for the model to recover from broken tracks using this representation since the model will then need to somehow fuse the end of one inputted pose with the beginning of a separate inputted pose.

To further help the model recover from broken tracks and unreliable pose keypoint estimation we also feed the entire RGB image as well into the model. This fusion approach of inputting both the rendered pose and the entire RGB image gives the model more opportunity to recover from any mistakes in the tracking and pose estimation by directly taking evidence from the RGB image itself.

Experiments with the \textit{ANV dataset} and \textit{Volleyball dataset} \cite{Ibrahim_2016_hierarchical_deep} demonstrate RePGARS's capability of performing group activity recognition in videos without the aid of reliable pose detection and tracking information. 

\begin{figure}[h]%
\centering
\includegraphics[width=1.0\textwidth]{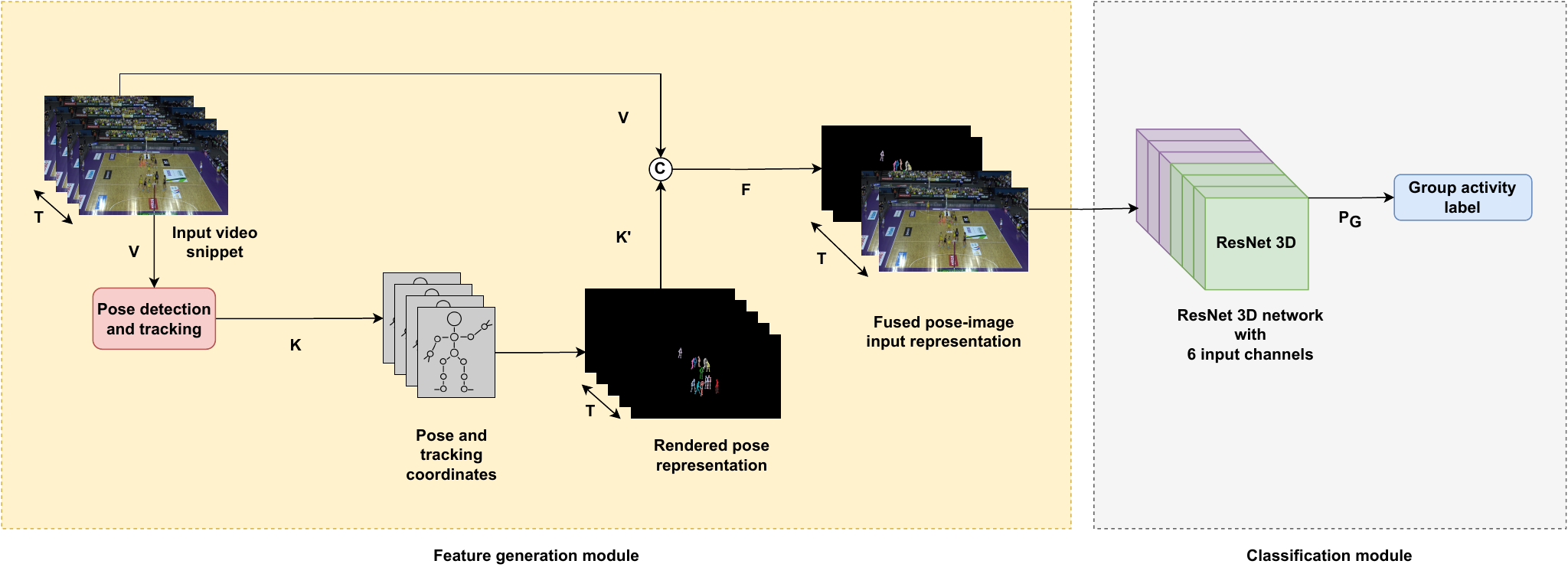}
\caption{Overview of Rendered Pose based Group Activity Recognition System (RePGARS).}
\label{f:RePGARS-model}
\end{figure}

\hyperref[f:RePGARS-model]{Figure~\ref*{f:RePGARS-model}} shows an overview of the proposed RePGARS approach. First, a pose detection and tracking algorithm is applied to extract pose keypoint associations of the individuals in the video. For this purpose, we employed the \textit{OpenPifPaf} real-time pose detection and tracking algorithm \cite{Kreiss2021_openpifpaf}. Predicted pose estimates are visually rendered against a black background, aligned with their corresponding frame-space position coordinates. Rendered pose representations are concatenated with the raw video stream to create a fused feature representation of the input. The concatenated features are subsequently input into a modified \textit{ResNet-3D} network \cite{hara2017_resnet3d_learning}, which is adapted to accommodate 6 input channels. This network predicts the group activity $p_G$. Detailed descriptions of each component of the proposed method are presented in the following sub-sections.

\subsection{Pose Detection and Tracking}
\label{ssec:pose-detection-tracking}

Since RePGARS is based on generating pose-image fused feature representation for input videos to learn group activities, the initial step of the process is human detection and tracking in the video and performing pose estimations of the respective individuals. 

There are many existing detection and 2D human pose estimation models which are appropriate for inferring such detection and pose keypoints. Human pose estimation algorithms come in two main variants. Pose estimation methods which first detect body joints and then group them to form individual poses per person are called bottom-up approaches. \textit{OpenPose} \cite{openpose}, \textit{OpenPifPaf} \cite{Kreiss2021_openpifpaf} and \textit{DeepCut} \cite{pishchulin2016_deepcut} are examples of prominent bottom-up pose detection and tracking algorithms. In contrast, top-down approaches first detect people in the scene and then predict keypoints for each detected person. \textit{Mask-RCNN} \cite{He_2017_mask_rcnn} is an example of a top-down pose estimation approach. 

Most of the prevailing pose-based group activity recognition approaches \cite{Gavrilyuk2020_ActorTransformer, Perez2020_SkeletonBased_GIRN, thilakarathne_2022_pogars} use manually annotated human tracking data coupled with top-down pose estimation approaches to retrieve pose information of the individuals from videos. This process provides relatively accurate pose keypoint estimation compared to using bottom-up approaches \cite{zheng2023_deep_pose_estimation}.

Despite the higher accuracy and global contextual understanding associated with top-down approaches in comparison to bottom-up approaches, the latter are often more suitable for real-time analytical applications due to lower computational complexity. Furthermore, bottom-up methods inherently handle occluded body parts and cluttered backgrounds by detecting joints independently of each other. This robustness ensures accurate pose estimation even in challenging visual conditions \cite{zheng2023_deep_pose_estimation}. 

Since bottom-up approaches are proven to better in real-time analytical scenarios with their lower computation complexity, We employ the \textit{OpenPifPaf} real-time pose detection and tracking algorithm \cite{Kreiss2021_openpifpaf} to extract pose information from videos. Our approach decouples tracking and pose estimation from the model used to classify the group activities. This enables the usage of any pose detection and tracking algorithm for the task of pose-image fused feature generation.

Given an input video clip $V$ spanning $T$ frames with $N$ people visible in each frame ($N$ can change across frames), \textit{OpenPifPaf} is used to estimate a set of corresponding tracked poses, $K$. For each individual, 17 2D pose keypoints are estimated, so $K\in \mathbb{R}^{N\times 34\times T}$ (assuming that each person is visible in all $T$ frames).

For the Volleyball and ANV datasets used in our experiments, we observed that the detections included non-player people, such as spectators and officials. Since these people are not directly involved in the group activities, we applied heuristics based on the court boundaries to remove them from the pose estimation data. The remaining tracked poses were then utilised to render the pose representations of each video. 

\subsection{Pose-image Fused Feature Representation}
\label{ssec:pose-image-fused-feature-representation}

Understanding the intricate spatio-temporal dynamics of individuals and their interactions in video is essential for accurate group activity recognition. Features derived solely from raw RGB data captured in video frames lack the emphasis on joint articulation and motion necessary to learn rich intermediate representations of complex human actions and interactions. To address this limitation, RePGARS leverages the pose detection and tracking estimations to create a fused feature representation that extends RGB data with explicit pose information.

\begin{figure}[h]%
\centering
\includegraphics[width=1.0\textwidth]{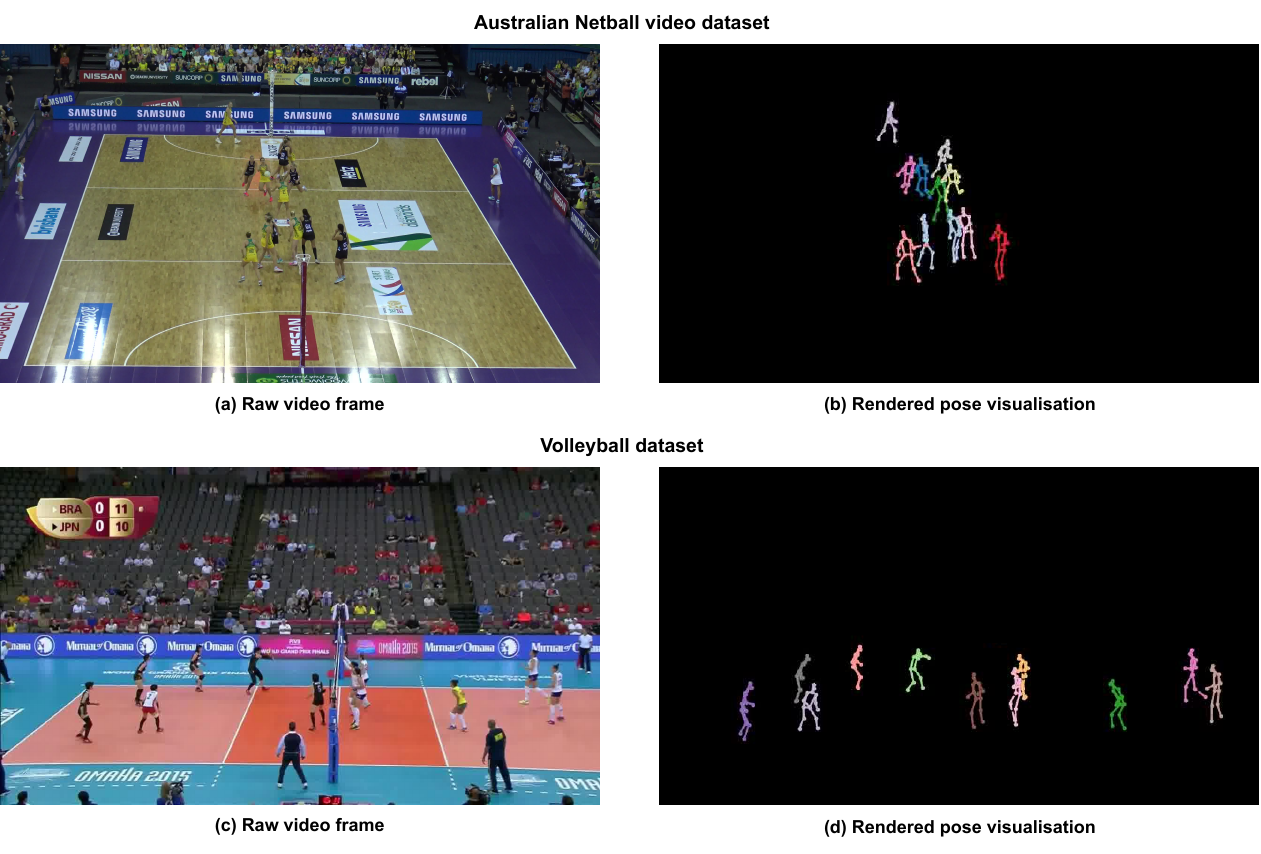}
\caption{Rendered pose visualisations of two sample video frames from ANV dataset and Volleyball dataset. Frame (a) is a sample from ANV dataset and frame (b) is its respective rendered pose visualisation. Frame (c) is a sample from Volleyball dataset and frame (d) is its respective rendered pose visualisation. Different colours have been used to denote each tracked individual in the video snippet.}
\label{f:rendered-pose-sample}
\end{figure}

For an RGB input video represented by 3 channels $V\in \mathbb{R}^{T\times 3\times H\times W}$ where $H$, $W$ denotes the height and width of the video snippet respectively: 

\begin{itemize}
  \item \textit{Step 1: } Given the pose detection coordinates \textit{$K$} of the input video \textit{$V$}, a new video snippet \textit{$K'\in \mathbb{R}^{T\times 3\times H\times W}$} is created by rendering the pose coordinates of the players within video frames against a black background (See \hyperref[f:rendered-pose-sample]{figure~\ref*{f:rendered-pose-sample}}). 
  
  \item \textit{Step 2: } The generated feature stream \textit{$K'$} is combined with the input video \textit{$V$}, by concatenating them along the channels dimension. This process results in the formation of a fused feature embedding denoted as \textit{$(V\mid\ K') = F \in \mathbb{R}^{T\times 6\times H\times W}$}.
\end{itemize}

Concatenated channels included in the pose-image fused feature representation \textit{$F$} represent pose information of individuals in the video snippet. Compared to existing group activity recognition methods \cite{Gavrilyuk2020_ActorTransformer, Perez2020_SkeletonBased_GIRN, thilakarathne_2022_pogars} which uses pose information as numerical input modalities, the proposed feature generation methodology of RePGARS has several advantages. 

Utilising established 3D CNN architectures in RePGARS offers a distinct advantage by making use of pretrained weights with transfer learning \cite{Yosinski2014_transferLearning} and training optimisation techniques.

Rendered pose provides a spatial representation, preserving the local context and relative positions of body parts of the individuals. Even without reliable tracking information, this visual representation offers sufficient information for the model to adapt to tracking errors and learn the spatial dynamics of the people present in the scene. Moreover, rendered pose representations can handle variations in player appearance and pose estimation that occur due to factors like lighting, occlusion, or clothing. As a result, the group activity recognition model is more likely to generalise to previously unseen data.

Interpretability and transparency of the predictive models are two important factors to consider in machine learning model development. Rendered visual representations leads to easy interpretability and visualisation of the pose information. In the context of RGB-based group activity recognition models, interpreting the importance of feature sets is challenging while in pose based methods researchers and practitioners can visually inspect rendered poses, aiding in understanding model predictions and potential errors.

In summary, pose-image fused feature generation provides a richer, spatially contextual, and visually interpretable representation of the input video, enhancing the RePGARS's robustness, adaptability, and effectiveness in group activity recognition.

\subsection{Spatio-temporal Evolution Modeling}
\label{ssec:spatio-temporal-evolution-modeling}

In RePGARS, the spatio-temporal dynamics of the scene are manifested by the fused feature embedding \textit{$F$}, explained in \hyperref[ssec:pose-image-fused-feature-representation]{section~\ref*{ssec:pose-image-fused-feature-representation}}. In order to form classification predictions based on the temporal evolution and spatial dynamics of videos represented by the feature representation, we employed a 3D CNN based network. 3D CNNs have the ability to learn hierarchical features representing complex motion patterns, which makes them ideal for group activity recognition in videos. We used a \textit{ResNet18-3D} network \cite{hara2017_resnet3d_learning} as the classification backbone in our experiments.

After the feature composition step, the enriched feature representation of the video is subsequently input into a modified \textit{ResNet-3D} network. The network is adapted to accommodate 6 input channels since \textit{$F\in \mathbb{R}^{T\times 6\times H\times W}$}. It predicts the group activity $p_G$. 

The \textit{ResNet-3D} model was initialised using publicly available weights pretrained on the \textit{Kinetics} dataset \cite{kay2017_kinetics}. Since RePGARS's fused feature representation is a combination of two visual cues (raw video stream and rendered pose representation stream), weights of the newly added channels were also set to pretrained values similar to the initial 3 channels. Having the capability to use transfer learning \cite{Yosinski2014_transferLearning} leads to improved performance and better generalisation of our group activity recognition approach. In order to train our proposed group activity recognition model, we used multi-class cross-entropy loss as the loss function.

\section{Experiments}
\label{sec:Experiments}

\subsection{Experimental Setup}
\label{ssec:experimental-setup}

RePGARS is evaluated on two sports activity datasets. The first dataset we used is the popular volleyball dataset \cite{Ibrahim_2016_hierarchical_deep}. It is the same dataset used for the experiments performed with POGARS \cite{thilakarathne_2022_pogars}. We also utilise the Australian Netball Video dataset, which is described in \hyperref[sec:anv-dataset]{section~\ref*{sec:anv-dataset}}.

We used the \textit{OpenPifPaf} real-time pose detection and tracking algorithm \cite{Kreiss2021_openpifpaf} to obtain bounding box tracklets and pose keypoint estimates for each player in videos of both datasets. OpenPifPaf is based on a \textit{Shufflenet-v2} \cite{ma2018_shufflenet} backbone and the network has been pretrained using the \textit{COCO} dataset \cite{COCO_Lin_2014} for predicting 17 keypoints \textit{(nose, eyes, ears, shoulders, elbows, wrists, hips, knees and ankles)}. 

The weights of the RePGARS classification model were optimised using \textit{ADAM} \cite{Kingma2014_adam} with initial learning rate set to $10^{-3}$, step-wise learning rate decay with a step size of 10 and fixed moving average decay rates $\beta_{1} = 0.9, \beta_{2} = 0.999$.
Training and evaluation were performed using the PyTorch deep learning framework \cite{paszke2019_pytorch} and an \textit{NVIDIA GeForce RTX 2080 Ti} GPU.

\subsection{Volleyball Dataset}
\label{ss:ch4-volleyball-dataset}

The volleyball dataset introduced by Ibrahim et al. \cite{Ibrahim_2016_hierarchical_deep} contains 55 videos collected from publicly available YouTube volleyball matches. 
It includes 4830 trimmed group activity instances which belong to 8 activity classes. 
Each activity instance contains 41 frames where the middle frame is labeled with the group activity label and individual action labels.
The 9 individual action labels are \textit{spiking, blocking, setting, jumping, digging, standing, falling, waiting,} and \textit{moving}. 
The 8 group activity labels are \textit{right spike, left spike, right set, left set, right pass, left pass, right winpoint,} and \textit{left winpoint}. In the experiments on RePGARS, we only use the group activity labels. From each video clip of the Volleyball dataset, 20 frames are utilised in model training and evaluation.

For the experiments we use the detection and pose estimates obtained using \cite{Kreiss2021_openpifpaf}. We have also followed the train/test splits suggested by \cite{Ibrahim_2016_hierarchical_deep}. In order to reduce model overfitting, we perform data augmentation by also training on horizontally flipped versions of examples from the training set and flipping the activity labels accordingly.

\subsection{Australian Netball Video Dataset}
\label{ss:ch4-netball-dataset}

The novel Australian Netball Video dataset, introduced in \hyperref[sec:anv-dataset]{section~\ref*{sec:anv-dataset}}, comprises 1801 annotations for instantaneous group events. To effectively train and assess deep learning models for group activity recognition, we define a single group activity instance as a span of 20 frames around the event annotation frame, considering the annotated event frame as the central reference point. 

Due to the limited number of annotations for \textit{gain} and \textit{turnover} events in the dataset, and compounded by their visual similarity in video streams, we opted for a simplified approach. Accordingly, we only utilised three distinct event classes: \textit{shot, goal circle feed,} and \textit{centre pass} to consider as group activity instances. These 3 activity classes are used to train and evaluate the group activity recognition model.

Similar to the procedure we followed with the volleyball dataset, detection and pose estimations of the ANV dataset are obtained using \cite{Kreiss2021_openpifpaf}. Among the 17 untrimmed videos within the dataset, 10 were allocated for model training purposes, while 4 were designated for validation, and the remaining 3 were utilised in the testing split. In order to reduce model overfitting, we perform data augmentation by also training on horizontally flipped versions of examples from the training set. 

\subsection{Experimental Results}
\label{sec:experiment-results}

\subsubsection{Ablation Study}
\label{sss:ablation-study}

A key element of RePGARS is the pose-image fused feature representation mechanism. The feature representation consists of rendered pose and raw RGB frames of a given video. We performed an extensive study in order to investigate the predictive power of these modalities and their combined strength. It comprises three distinct settings. In the initial configuration, the input for the classification model consists solely of RGB features extracted from the video frames. In the second setting, the input is derived from rendered pose features, elaborated in \hyperref[ssec:pose-image-fused-feature-representation]{section~\ref*{ssec:pose-image-fused-feature-representation}}. Lastly, the third setting combines both RGB information and rendered pose information, as depicted in \hyperref[f:RePGARS-model]{figure~\ref*{f:RePGARS-model}}, creating a fused feature setting for the classification model. We performed the experiments with the volleyball dataset \cite{Ibrahim_2016_hierarchical_deep} and ANV dataset to assess the impact of these settings on overall performance of the proposed group activity recognition model. 

\begin{table}[h]
\small
\caption{Comparison of validation accuracy with different feature composition settings of RePARS evaluated on Volleyball dataset and ANV dataset.}
\label{tab:ablation-study}
\begin{tabular}{lcc}
\toprule
\multirow{2}{*}{Method}     & \multicolumn{2}{c}{Accuracy}                                                                                                 \\  
                            & \begin{tabular}[c]{@{}c@{}}Volleyball dataset\end{tabular} & \begin{tabular}[c]{@{}c@{}}ANV dataset\end{tabular} \\ \hline
RePGARS  - with only RGB features                 & 80.2                                                         & 73.1                                                          \\
RePGARS  - with only rendered pose features       & 85.0                                                         & 75.1                                                          \\
\textbf{RePGARS ALL - with pose-image fused features} & \textbf{86.8}                                                         & \textbf{78.4}                                                          \\ \botrule
\end{tabular}
\end{table}

The results in \hyperref[tab:ablation-study]{table~\ref*{tab:ablation-study}} show, RePGARS using only rendered pose features performs significantly better than RGB features, highlighting the importance of pose information in group activity recognition. Pose-image fused feature representation which combines the pose information with RGB features achieves the highest accuracy for group activity recognition in both datasets. RePGARS ALL outperforms its counterpart approach which only uses RGB features by 6.6\%  and 5.3\% on the volleyball dataset and ANV dataset respectively. This demonstrates the effectiveness of utilising fused pose-image feature representation to learn spatio-temporal dynamics in group activities.  

\subsubsection{Performance Comparison of RePGARS Vs. POGARS}
\label{sss:ch4-performance-comparison-pogars}

The key advantage of RePGARS over existing group activity recognition techniques lies in its capacity to effectively incorporate imperfect detection and tracking data for capturing the spatio-temporal behaviours of individuals in video footage. To validate this hypothesis, we conducted experiments comparing the performance of RePGARS with POGARS \cite{thilakarathne_2022_pogars} under conditions where accurate detection and tracking information were available and in situations where reliable tracking details were absent. \hyperref[tab:pogars-vs-repgars]{Table~\ref*{tab:pogars-vs-repgars}} summarises the results from experiments conducted on two datasets: the volleyball dataset and ANV dataset.

During the experiments, we explored two different person level fusion approaches with POGARS. In early person fusion, person-level feature independence is removed before modeling the temporal evolution by fusing the tracked pose features of all people before input into the module that captures temporal dynamics. Alternatively, in late person fusion the temporal evolution of each person is first modeled individually before the learned features are fused together at the end.

In the experiments conducted using the volleyball dataset, we acquired manually annotated ground-truth detection and tracking data for individuals from \cite{Sendo2019_HeatmappingOfPeople}. Pose details of individuals were extracted utilising the \textit{Stacked Hourglass} pose estimation algorithm \cite{Newell_2016_stackedHourGlass}. This particular detection method is referenced as "GT" in the \hyperref[tab:pogars-vs-repgars]{table~\ref*{tab:pogars-vs-repgars}}. Subsequently, the \textit{OpenPifPaf} real-time pose detection and tracking algorithm \cite{Kreiss2021_openpifpaf} was employed to obtain tracking and pose data from both datasets, representing the less reliable detection and tracking information, as depicted in \hyperref[f:track-fails]{figure~\ref*{f:track-fails}}.

\begin{table}[h]
\caption{Comparison of group activity recognition accuracy between RePGARS and POGARS with different detection and tracking approaches. The results are partitioned into two categories. The first category contains methods that use ground truth detection and tracking. The second category contains methods that rely on \textit{OpenPifPaf} detection and tracking algorithm for group activity recognition.}
\label{tab:pogars-vs-repgars}
\begin{tabular}{lccc}
\toprule
\multirow{3}{*}{Method}              & \multirow{3}{*}{\begin{tabular}[c]{@{}c@{}}Detection and \\ tracking approach\end{tabular}} & \multicolumn{2}{c}{Accuracy}                                                                                                 \\ 
                                     &                                                                                    & \begin{tabular}[c]{@{}c@{}}Volleyball\\ dataset\end{tabular} & \begin{tabular}[c]{@{}c@{}}ANV\\ dataset\end{tabular} \\ \hline
POGARS - Early person fusion         & GT                                                                                 & 83.0                                                         & -                                                             \\
POGARS - Late person fusion & GT                                                                       & \textbf{88.3}                                                & -                                                    \\ \smallskip 
RePGARS         & GT                                                                                 & 87.9                                                         & -                                                             \\ 
\hdashline 
\addlinespace

POGARS - Early person fusion         & OpenPifPaf                                                                         & 70.8                                                         & 51.0                                                          \\
POGARS - Late person fusion          & OpenPifPaf                                                                         & 74.0                                                         & 52.1                                                          \\\smallskip
RePGARS                     & OpenPifPaf                                                                & \textbf{86.8 }                                               & \textbf{78.4 }                                                \\

 \hline
\end{tabular}
{\footnotesize 
"GT" stands for ground-truth player position coordinates of volleyball dataset \cite{Ibrahim_2016_hierarchical_deep} obtained from \cite{Sendo2019_HeatmappingOfPeople}.
}
\end{table}

When ground truth detection and tracking was used POGARS with late person fusion only slightly outperforms RePGARS (88.3\% compared to 87.9\%) on the volleyball dataset. This means the rendered pose representation used by RePGARS is highly competitive compared to POGARS even when ground truth detection and tracking was used, although RePGARS was designed to be tolerant of imperfect pose.

When the automated tracked pose algorithm OpenPifPaf was used, RePGARS significantly outperforms both early and late fusion POGARS in both the volleyball and ANV datasets. This is a really important result since it is a test of the more realistic situation where unreliable tracking is used. Furthermore the performance of RePGARS degrades by only 1.1\% for the volleyball dataset when unreliable tracked pose is used instead of ground truth tracking. In contrast, POGARS performance degrades by 14.3\% when unreliable tracked pose information is used. This shows using the rendered pose representation is in deed more tolerant to poor quality tracking compared to the keypoint pose representation used by POGARS.

The accuracy of POGARS using early fusion drops by 12.2 percentage points, while the late fusion approach drops by 14.3 percentage points for the volleyball dataset. This shows late fusion is more affected by poor tracking information and has a larger degradation of performance compared to early fusion due to late fusion being more effected by broken tracks. The drop in the performance of early fusion is due to poor quality of the pose extract by OpenPifPaf compared to the pose extracted by ground truth detections.

POGARS achieves only 52.1\% accuracy on the ANV dataset compared to the accuracy of 78.4\% achieved by RePGARS. This shows the large difference in accuracy between POGARS versus RePGARS when a new dataset is introduced with no corresponding hand annotation. These findings underscore the robustness of RePGARS in handling less reliable detection and tracking information compared to POGARS.

\subsubsection{Performance Comparison of RePGARS Against Existing Approaches}
\label{sss:performance-comparison}

We compare the performance of RePGARS with state-of-the-art methods and 2 baselines. \hyperref[tab:repgars-comparison]{Table~\ref*{tab:repgars-comparison}} reports the group activity recognition accuracy of the selected models in the volleyball dataset \cite{Ibrahim_2016_hierarchical_deep}.

The baselines are: 
\begin{itemize}
  \item \textbf{Baseline 1 - Activity recognition with key-frame spatial features:} This baseline uses the middle frame of the activity video clip (21\textsuperscript{st} frame among 41 frames) as the input to a \textit{ImageNet} pretrained \textit{ResNet34} model \cite{He_2016_resent}.
  \item \textbf{Baseline 2 - Activity recognition with spatio-temporal RGB features:} This baseline takes 20 consecutive frames of the collective activity as input to a network based on \textit{ResNet18} architecture with 3D CNN layers. baseline 2 can be thought of as a temporally extended version of baseline 1.
\end{itemize}

\begin{table}[h]
  \caption{Comparison of different baselines and state-of-the-art methods with RePGARS evaluated on volleyball dataset. The results are partitioned into two categories. The first category contains methods that do not use any ground truth annotations at all. The second category contains methods that rely on ground truth pose and/or ground truth tracking, which are often not available in real-world scenarios. }
  \label{tab:repgars-comparison}
\begin{tabular}{lcc}
\toprule
Method & Model input & Accuracy \\
\midrule
Baseline 1 - Keyframe based model                & RGB & 73.3 \\
Baseline 2 - Spatio-temporal feature based model & RGB & 80.2 \\
Two-stage Hierarchical model
\cite{Ibrahim_2016_hierarchical_deep}            & RGB & 81.9 \\
CERN \cite{CERN}                                 & RGB & 83.3 \\
I3D \cite{kay2017_kinetics}                      & RGB & 84.6 \\ \smallskip
\textbf{RePGARS}                                 & \textbf{RGB + Pose} & \textbf{86.8} \\
\hdashline
\addlinespace
Multi-stream CNN \cite{Azar2018_multistream}     & GT tracking + RGB + Optical flow  & 90.5 \\
Spatio-temporal attention based model
\cite{Lu2019_spatioTempAtt}                      & GT tracking + RGB + Pose       & 91.7 \\
Convolutional Relational Machine
\cite{Azar2019_convoRelationalMachine}           & GT tracking + RGB + Optical flow  & 93.0 \\
POGARS \cite{thilakarathne_2022_pogars}          & GT tracking + Pose             & 88.3 \\
POGARS ALL \cite{thilakarathne_2022_pogars}      & GT tracking + Pose             & 93.2 \\
\textbf{POGARS with ball}
\cite{thilakarathne_2022_pogars}                 & \textbf{GT tracking + Pose + Ball} & \textbf{93.9} \\
\bottomrule
\end{tabular}
\end{table}

The results in \hyperref[tab:repgars-comparison]{Table~\ref*{tab:repgars-comparison}} are divided into two halves (below and above the dotted line). The results in the first half are from models that do not use any ground truth annotations. In contrast the results in the second half show the results for methods that use some kind of ground truth information whether it is ground truth tracked pose or ground truth tracking. The most significant result from the table is the fact that RePGARS is the best performing model among all methods that do not use any ground truth annotations. This is due to RePGARS' ability to take advantage of unreliable pose information.

Baseline 1 employs keyframe-based spatial feature extraction to analyse group activities. Despite not accounting for the temporal aspects of these activities, the model demonstrates basic proficiency in recognising collective activities by utilising state-of-the-art image classification network and transfer learning \cite{Yosinski2014_transferLearning}.

Compared to baseline 1, baseline 2 requires more memory and computational power because of its use of 3D convolutional layers in the network. The 6.9\% accuracy improvement observed in baseline 2 over baseline 1 highlights the effectiveness of 3D CNNs in capturing motion information for recognising group activities. Similarly, in our proposed group activity recognition system, we employ 3D CNNs to extract spatio-temporal features from videos. Compared to these baseline approaches RePGARS reports a significantly higher achieved accuracy as well as faster model convergence.

The study by \cite{Azar2019_convoRelationalMachine} achieved a 6.2\% increase in accuracy compared to RePGARS by using optical flow and incorporating ground truth tracking which is impractical in the real world. The primary factor contributing to effective performance of the method is the utilisation of reliable tracking information for individuals within the scene. In addition, optical flow faces practical challenges due to its computational intensity, sensitivity to noise, and assumptions like brightness constancy. These limitations impact its accuracy in real-world scenarios, particularly in dynamic environments with complex motion patterns and resource constraints \cite{fortun2015optical}. It highlights the efficiency and effectiveness of RePGARS in practical scenarios.

To the best of our knowledge, RePGARS stands as a unique approach employing real-time pose detection and tracking for group activity recognition, distinguishing itself from other methods utilising ground truth tracking information. Despite relying on detection and tracking information prone to occasional inaccuracies, RePGARS exhibits commendable performance when compared to methods utilising ground-truth tracking pose information (See \hyperref[tab:repgars-comparison]{table~\ref{tab:repgars-comparison}}).

Notably, the difference in accuracy between RePGARS and POGARS \cite{thilakarathne_2022_pogars} is a mere 1.5\%. POGARS benefits from precise ground truth pose and tracking data, whereas RePGARS relies on pose detection and tracking generated in real-time through the \textit{OpenPifPaf} algorithm \cite{Kreiss2021_openpifpaf}. This marginal difference emphasises the effectiveness of RePGARS, highlighting its potential even in the absence of perfect tracking information.

\begin{figure}[ht!]%
\centering
\includegraphics[width=1.0\textwidth]{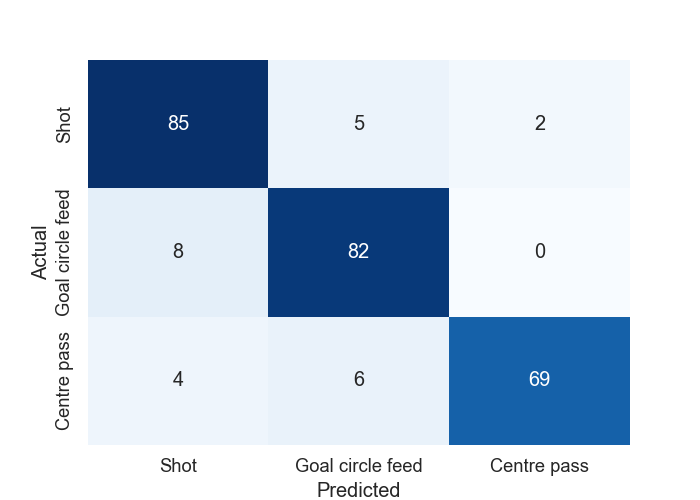}
\caption{Confusion matrix obtained using best performing RePGARS for ANV dataset.}
\label{f:netball-confusion-matrix}
\end{figure}

\hyperref[f:netball-confusion-matrix]{Figure~\ref{f:netball-confusion-matrix}} and \hyperref[f:volleyball-confusion-matrix]{Figure~\ref{f:volleyball-confusion-matrix}} contains confusion matrices summarising correct and incorrect group activity label predictions on ANV dataset and Volleyball dataset respectively. 

The confusion matrix in \hyperref[f:netball-confusion-matrix]{Figure~\ref*{f:netball-confusion-matrix}}demonstrates RePGARS has remarkable predictive accuracy across all three classes in the ANV dataset. The confusion between \textit{"goal centre feed"} and \textit{"shot"} activities can likely be attributed to the spatial proximity of these activities on the netball court, often leading to similar player configurations.

\begin{figure}[ht!]%
\centering
\includegraphics[width=1.0\textwidth]{volleyball_confusion_matrix.png}
\caption{Confusion matrix obtained using best performing RePGARS for Volleyball dataset.}
\label{f:volleyball-confusion-matrix}
\end{figure}

The confusion matrix for the volleyball dataset \hyperref[f:volleyball-confusion-matrix]{Figure~\ref*{f:volleyball-confusion-matrix}} shows a similar pattern. While performing well in most of the group activity classifications, RePGARS has predicted occasional misclassifications between \textit{"pass"} and \textit{"set"} activities. This confusion is rooted in the similar player configurations during these activities, underscoring the nuanced challenges faced in accurately distinguishing them. 

\section{Conclusion}
\label{sec:Conclusion}

In this study, we introduced RePGARS, an innovative framework designed to identify group activities in sports videos. By employing a unique fused feature representation derived from rendered pose data, RePGARS classifies group activities in videos. Our experiments show RePGARS stands out as the best performing method for group activity recognition, when the model is not provided with any ground truth annotations as input.

Despite utilising unreliable tracked pose information, RePGARS achieves impressive accuracy in recognising group activities, especially notable in the context of the Volleyball dataset, dropping accuracy by only 1.1\% compared to using ground truth tracking when the previous state-of-the-art POGARS drops accuracy by 14.3\%. Therefore this approach is particularly advantageous in real world scenarios where tracking information and keypoint estimation is unreliable.

Additionally, we conducted experiments on the newly introduced Australian Netball Video dataset, specifically tailored for sports group activities. Our findings emphasise RePGARS' effectiveness in the group activity recognition domain. Looking ahead, our future research aims to address the challenge of instantaneous event detection in untrimmed videos, utilising unreliable tracking and detection information as the input modalities.

\section*{Acknowledgement}
\label{sec:Acknowledgement}

This research was supported supported by Australian Institute of Sport and Netball Australia. We are thankful to Dr. Mitch Mooney who provided the domain expertise in Netball that greatly assisted the research.

\section*{Data Availability}
\label{sec:data-availability}

\begin{itemize}
    \item The Volleyball dataset is available at \cite{Ibrahim_2016_hierarchical_deep}.
    \item Australian Netball Video dataset that support the findings of this study are available from the Netball Australia. Restrictions apply to the availability of these data, which were used under license for the current study and so are not publicly available. However, data are available from the authors upon reasonable request and with permission from the Netball Australia.
\end{itemize}

\bibliography{sn-bibliography}

\end{document}